\documentclass[sigconf]{acmart}

\usepackage[utf8]{inputenc}
\usepackage[T1]{fontenc}

\usepackage{mathtools}
\usepackage{amsfonts}
\usepackage{etoolbox,siunitx}
\robustify\bfseries
\usepackage{booktabs}
\usepackage{multirow}

\usepackage{caption}
\usepackage{subcaption}

\usepackage{hyperref}
\usepackage[capitalise,noabbrev]{cleveref}

\newcommand{\algoo}{SuSL4TS}

\begin{document}
\title{Semi-unsupervised Learning for Time Series Classification}
\author{Padraig Davidson}
\email{davidson@informatik.uni-wuerzburg.de}
\affiliation{
\department{Chair of Computer Science X}
\institution{University of Würzburg}
\city{Würzburg}
\country{Germany}
}
\author{Michael Steininger}
\email{steininger@informatik.uni-wuerzburg.de}
\affiliation{
\department{Chair of Computer Science X}
\institution{University of Würzburg}
\city{Würzburg}
\country{Germany}
}
\author{André Huhn}
\email{andre.huhn@stud-mail.uni-wuerzburg.de}
\affiliation{
\department{Chair of Computer Science X}
\institution{University of Würzburg}
\city{Würzburg}
\country{Germany}
}
\author{Anna Krause}
\email{anna.krause@informatik.uni-wuerzburg.de}
\affiliation{
\department{Chair of Computer Science X}
\institution{University of Würzburg}
\city{Würzburg}
\country{Germany}
}
\author{Andreas Hotho}
\email{hotho@informatik.uni-wuerzburg.de}
\affiliation{
\department{Chair of Computer Science X}
\institution{University of Würzburg}
\city{Würzburg}
\country{Germany}
}

\begin{abstract}
Time series are ubiquitous and therefore inherently hard to analyze and ultimately to label or cluster.
With the rise of the Internet of Things (IoT) and its smart devices, data is collected in large amounts any given second.
The collected data is rich in information, as one can detect accidents (e.g. cars) in real time, or assess injury/sickness over a given time span (e.g. health devices).
Due to its chaotic nature and massive amounts of datapoints, timeseries are hard to label manually.
Furthermore new classes within the data could emerge over time (contrary to e.g. handwritten digits), which would require relabeling the data.

In this paper we present \textit{\algoo}, a deep generative Gaussian mixture model for semi-unsupervised learning, to classify time series data.
With our approach we can alleviate manual labeling steps, since we can detect sparsely labeled classes (semi-supervised) and identify emerging classes hidden in the data (unsupervised).
We demonstrate the efficacy of our approach with established time series classification datasets from different domains.
\end{abstract}



\maketitle

\section{Introduction}
Autoencoders (AE) have become the de-facto standard for anomaly detection within deep learning.
In this pair of neural networks, one trains an encoder to map the features of the input data (i.e. time series) into the latent space.
The decoder reconstructs this representation as well as possible, with the constraint that the latent space is much smaller than the input domain.
After training on data without anomalies (e.g. normal data), predictions of anomalies can be done by defining a threshold on the anomaly score (e.g. reconstruction loss) to predict abnormalities.
In recent years, variational AEs~\cite{kingma2013auto} (VAE), have gained popularity, since they encode the data distribution in the latent space, rather than the raw features.
This allows training on all variations of data and thus reliefs the burden of filtering data beforehand.
Furthermore one can see anomaly detection as a probability rather than a raw score~\cite{an2015variational}.

Since time series are ubiquitous and present in a myriad of types for classification, we are interested in models beyond this binary classification task.
With the development of semi-supervised generative models~\cite{kingma2014semi}, we are able to classify time series data, while only having to label a smaller amount of data.
But, we still need to know all manifestations of classes beforehand.
On the other hand, we could cluster the data, needing no label information at all~\cite{aghabozorgi2015time}.
This however, often comes with the drawback of lower classification accuracy and the need to manually annotate the found clusters.

\begin{figure*}
    \centering
    \includegraphics{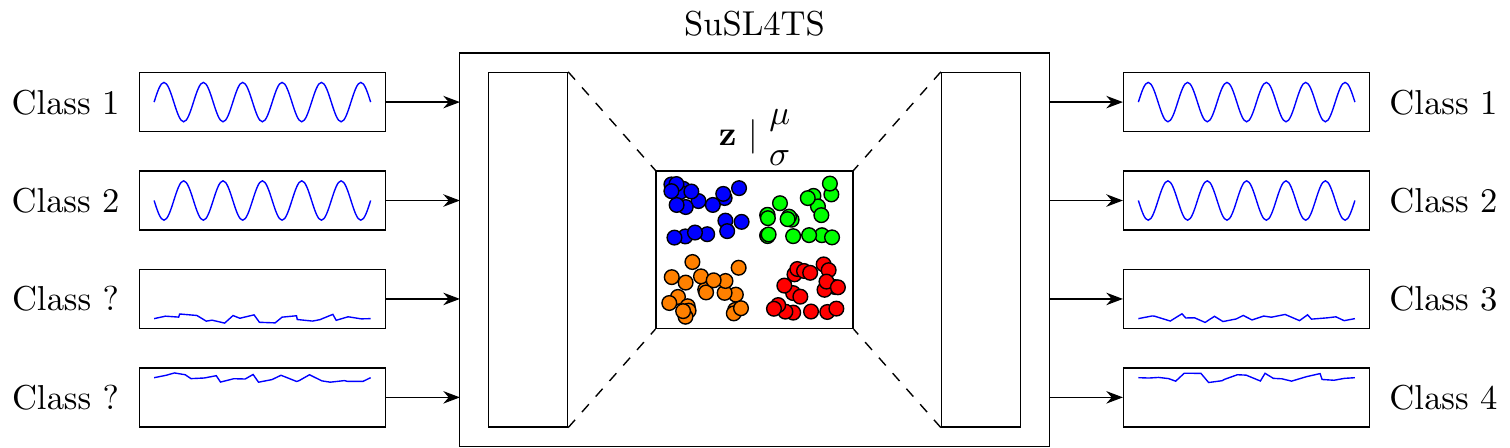}
    \caption{Semi-unsupervised learning for time series data (SuSL4TS).
    The model is tasked to classify the data on the left hand side (a single multivariate time series), with only limited labels available for classes 1 and 2, and two completely unknown classes.
    The output on the right hand side is the classified data with all four classes found.
    Within the model we can see four distinct cluster automatically found when zooming in on the latent space.}
  \label{fig:teaser}
\end{figure*}

To combine the benefits of the high classification accuracy in semi-supervised models with the ability to detect new classes, the hybrid approach of \textit{semi-unsupervised} learning~\cite{davidson2021semi,willetts2020semi} has emerged.
In this paper we present \textit{\algoo}, a convolutional Gaussian mixture model for semi-unsupervised learning on time series data.
\Cref{fig:teaser} visualizes the basic principle of our approach.
Our contributions are twofold:
(1) We present a model capable of semi-unsupervised time series classification from raw time series, partially on par with state of the art models, while only needing a limited amount of labels,
(2) We show the efficacy of our approach on several benchmark datasets, and perform extensive experiments in this new domain for time series.

The remainder of this paper is structured as follows: after presenting related work in the field, we present the used datasets in~\cref{sec:datasets}.
\Cref{sec:methodology} illustrates the foundations of the used model, while \cref{sec:experiments} outlines our experiments.
We conclude with a discussion (\cref{sec:discussion}) of the experiments and depict future work in~\cref{sec:summary}.

\section{Related Work}
\label{sec:related_work}
Related work in time series classification is manifold since most datasets and solutions are customary.
A larger review and benchmark of different algorithms and datasets can be found in~\cite{bagnall2017great,ruiz2021great}.
The authors present algorithms suited for time series classification task in an univariate~\cite{bagnall2017great} and a multivariate~\cite{ruiz2021great} setting.
Univariate refers to datasets in which only a single sensor is used for the classification, whereas multivariate considers multiple sensor readings at the same time.
The best performing algorithms are often \textit{BOSS}~\cite{schafer2015boss}, \textit{COTE}~\cite{lines2018time} and \textit{TS-Chief}~\cite{shifaz2020ts} in the univariate problems, whereas \textit{ROCKET}~\cite{dempster2020rocket}, \textit{WEASEL}~\cite{schafer2017multivariate} and \textit{INCEPT-NET}~\cite{ismail2020inceptiontime} show great performance in the multivariate problems.
We refer the reader to the mentioned papers for a detailed explanation of the algorithms and specific datasets.
A systemic analysis of related work can be found in~\cite{lucas2019proximity}, in which the authors list time series classifiers based on their technique: distance based, features based, ensemble approaches and tree approaches.
In general, most algorithms aim for fully supervised classification, whereas we aim to reduce the time spent for labeling datasets.

Since the field of semi-unsupervised learning is relatively new, there is limited related work in this domain.
In~\cite{willetts2018semi}, the authors present a network capable of semi-unsupervised time series classification on human activity recognition.
They compare their approach to the semi-supervised M2-model~\cite{kingma2014semi}, and show great performance even with classes hidden.
Their approach uses extracted features from the time series and use them within the fully connected network.
We are however interested in the use of the raw time series signals, and therefore focus on the convolutional approach presented in~\cite{davidson2021semi}.

\section{Datasets}
\label{sec:datasets}
The following datasets were used for the experiments.
Since we are especially interested in the learning paradigms of semi-supervised and semi-unsupervised learning, we only hand-select some datasets for our purposes.
The main strength of our approach are most prominent, if we have a large dataset that would require huge efforts to label, and only a limited amount of known classes in comparison to the whole dataset.
Therefore we use three datasets to perform our experiments, both in the univariate and multivariate setting, stemming from different domains of data acquisition, which meet our requirements.
The datasets are introduced in the following.
A short tabular view is available in~\cref{tab:datasets}.
\begin{table*}
    \caption{Datasets used for the experiments}
    \label{tab:datasets}
    \centering
    \begin{tabular}{@{}lllllll@{}}
    \toprule
    Dataset & Features & Classes & Samples from & Sampling rate & Input size & Best accuracy\\
    \midrule
    HAR & accelerometer \& gyroscope data & 6 & \num{30} subjects & \SI{50}{\hertz} & $\mathbb{R}^{9 \times 128}$ & \SI{96}{\percent} (Multiclass SVM~\cite{anguita2013public})\\
    ECG & electrocardiogram recording & 5 & \num{47} subjects & \SI{360}{\hertz} & $\mathbb{R}^{1 \times 186}$ & \SI{93}{\percent} (CNN~\cite{Kachuee_2018})\\
    El. Devices & electrical consumption & 7 &  \num{251} households & every \SI{2}{\minute} & $\mathbb{R}^{1 \times 96}$ & \SI{80}{\percent} (BOSS~\cite{shifaz2020ts})\\
    \bottomrule
    \end{tabular}
\end{table*}

\subsection{Human Activity Recognition}
The Human Activity Recognition (HAR) dataset~\cite{anguita2013public} consists of data collected from accelerometer and gyroscope sensors in smartphones.
The subjects (\num{30}), aged 19 to 48, were tasked with performing various Activities of Daily Living (ADL) while carrying a smartphone on their waist.
The subjects were instructed to perform six distinct ADL adhering to a defined protocol outlining the order of activities.
The selected activities were \textit{standing, sitting, laying down, walking, walking downstairs} and \textit{walking upstairs}.
Each activity was performed for \SI{15}{\second}, except walking up- and downstairs which only lasted \SI{12}{\second}.
Each activity was performed twice throughout the routine and \SI{5}{\second} pauses separated activities.
Linear acceleration and angular velocity in three axes was recorded at a sampling rate of \SI{50}{\hertz}.
The data was then pre-processed for noise reduction.
Additionally gravitational and body motion was separated using a low-pass filter.
A total of 9 signals were sampled with a window of \SI{2.56}{\second} with \SI{50}{\percent} overlap (i.e. input is of size $\mathbb{R}^{9 \times 128}$).
A feature vector was obtained from each sampling window.
A total of 561 features were extracted with measures common in HAR literature like mean, correlation, signal magnitude area and autoregression coefficients as well as energy of different frequency bands, frequency skewness and the angle between vectors like mean body acceleration and the \textit{y} vector.
The data was randomly divided into a 70/30 training/test split containing \num{7352} and \num{2947} samples respectively as shown in~\cref{tab:har_classes}.
The authors also used a multiclass SVM to achieve a \SI{96}{\percent} classification accuracy on the dataset~\cite{anguita2013public}.

\subsection{ECG Heartbeat Classification}
The second dataset used was the MIT-BIH Arrythmia Dataset~\cite{Kachuee_2018}.
It consists of electrocardiogram (ECG) recordings from \num{47} subjects recorded at a \SI{360}{\hertz} sampling rate.
The recordings are grouped in five categories based on annotations by cardiologists.
As can be seen in~\cref{tab:ecg_classes} the class frequency is skewed towards the N class, which is to be expected since it includes normal heart behavior.
Further explanation of the individual classes can be found in~\cite{Kachuee_2018}.
Furthermore each entry in the set consists of a single heartbeat padded with zeroes to ensure consistent length (i.e. input is of size $\mathbb{R}^{1 \times 186}$).
The authors of~\cite{Kachuee_2018} propose a CNN architecture to classify the dataset in a fully supervised fashion, achieving an accuracy of \SI{93}{\percent}.

\subsection{Electric Devices}
The Electric Devices dataset contains measurements from households observing their electrical consumption.
Samples are taken every \SI{2}{\minute} from \num{251} households.
After pre-processing and resampling to \SI{15}{\minute} averages, the samples have a length of \num{96} values (i.e. input is of size $\mathbb{R}^{1 \times 96}$).
We use the version from~\cite{bagnall2012transformation}\footnote{Downloaded at \url{https://timeseriesclassification.com}}, regrouping the originally ten classes to seven: kettle; immersion heater; washing machine; cold group; oven/cooker; screen group and dishwasher.
The authors of~\cite{shifaz2020ts} report the best accuracy by using BOSS at \SI{80}{\percent}.

\section{Methodology}
\label{sec:methodology}
\paragraph{Gaussian Mixture Models}
A variational autoencoder in general consists of an encoder $\Phi(x): \mathbb{R}^n \rightarrow \mathbb{R}^d$, mapping the input data of dimensions $n$ into the latent space of dimensions $d$~\cite{kingma2013auto}.
The decoder $\Theta(z): \mathbb{R}^d \rightarrow \mathbb{R}^n$ on the other hands inverts this mapping, recreating the input data from the (compressed) latent encoding.
This compressed space is often used for other downstream tasks in a second training step, for example classification or other information extraction tasks.
The first step in this process is trained unsupervised, as it requires no annotated data, whereas the latter task is trained fully supervised.

This two-step process can be merged into one, by adapting the joint probability distribution $p_\Theta$, resulting in a Gaussian Mixture Deep Generative model (GMM) capable of learning semi-supervised classification~\cite{kingma2014semi}.
With some further modification we can use the inductive bias requirement to perform semi-unsupervised classification tasks with GMMs~\cite{davidson2021semi,willetts2020semi,willetts2018semi}.

\paragraph{GMM for Semi-unsupervised Classification}
In this work, we built on the work presented in~\cite{davidson2021semi,willetts2020semi} and adapt it to perform time series classification on raw sensor signals.
That is, we are interested in the improved pattern recognition and performance shown in~\cite{davidson2021semi}.
Therefore we adapt their work and replace the 2d convolutional networks for image classification, with 1d convolutions for raw time series.
Additionally we use the work shown in~\cite{willetts2020semi} in two ways.
Firstly, we use it a reference in performance for the presented convolutional model.
Secondly, we adapt their idea of the Gaussian $L_2$ regularization with the standard $L_2$ term provided by the Adam optimizer~\cite{kingma2014adam}.
Our overall loss function can now be described as~\cite{davidson2021semi}
\begin{align*}
\mathcal{L} &:= \underset{x,y \in D_l}{\mathbb{E}} \left[\mathcal{L}_l(x, y) - \alpha \cdot\log q_\Phi(y|x)\right] \\
    &+ \underset{x \in D_u}{\mathbb{E}} \left[\mathcal{L}_u(x) - \gamma \cdot \lambda \cdot \sum_{c \in C} q_\Phi(c | x) \cdot \log q_\Phi(c | x)\right] \\
    &+ w \cdot \Theta_t.
\end{align*}
$D_l$ refers to the labeled subset of the data, thus containing samples $x$ and their corresponding class $y$ (one-hot encoded).
On the other hand, $D_u$ contains all unlabeled data.
That is all data that is to be mapped to the known classes (semi-supervised classification), and data stemming from possibly new classes (unsupervised clustering).
$\Theta_t$ holds the trainable weights at epoch $t$.
$\alpha, \gamma, \lambda$ are hyperparameters weighting the entropy regularization, whereas the loss terms $\mathcal{L}_l, \mathcal{L}_u$ measure the evidence lower bound (ELBO) from the GMM model.
All other loss terms, the network architecture and further details can be seen in~\cite{davidson2021semi}.
This approach allows us to analyze our experiments in four learning regimes: unsupervised, semi-unsupervised, semi-supervised and fully supervised.

\section{Experiments}
\label{sec:experiments}
\paragraph{Experimental Setup}
Since \algoo\ is capable of handling all learning paradigms, we have conducted experiments in any setting.
That is, we performed a parameter search for \num{60} trials in the settings unsupervised learning (UL, \SI{0}{\percent} labeled), semi-supervised (SSL, \SI{20}{\percent} and \SI{50}{\percent} labeled), semi-unsupervised (SuSL, \SI{20}{\percent} and \SI{50}{\percent} labeled) and supervised (SL, \SI{100}{\percent} labeled).
In the semi-unsupervised setting, we hid different classes.
For the HAR dataset we tested three settings: 
(1) hiding all \textit{walking} classes,
(2) hiding all stationary classes, and
(3) hiding one movement (walking) and one stationary (laying) class,
while using the remainder semi-supervised.
For the ECG dataset we used two hiding schemes:
(1) omitting all normal heart beats, and
(2) omitting classes Q and V.
Within the electric devices dataset, we used two settings:
(1) hiding classes 1--3, and
(2) hiding classes 4--7. 
This was chosen arbitrarily, since there is no inherent split from the data\footnote{Classes are only labeled 1--7, while the mapping to the named version is missing.}.
When hiding classes with different sizes, we used \SIlist{20;50}{\percent} of each class.
That is, we did not use subsampling.
Finally, each time series was scaled via standard scaling/z-normalization.

For all settings and datasets we tested different types of networks.
The first one is the fully convolutional approach, using a convolutional feature extractor and decoder (\algoo).
In the second setting, we tested a fully connected model, using MLPs in the encoder, as well as the decoder (MLP SuSL).
We also experimented with a mixture (convolutional encoder and linear decoder) but found its performance to consistently lay between the other two settings.

\paragraph{Implementation}
All networks were implemented using the PyTorch~\cite{paszke2019pytorch} framework.
We chose the Adam optimizer~\cite{kingma2014adam} for training and performed a Bayesian hyperparameter search using optuna~\cite{akiba2019optuna} for each learning paradigm and dataset.
The search space for each parameter can be seen in~\cref{tab:experiments}.
\begin{table}[t]
    \caption{Hyperparameter search spaces.
    Optimization is done with \textit{optuna}~\cite{akiba2019optuna} for \num{60} runs in each experiment.
    }
    \label{tab:experiments}
    \centering
    \begin{tabular}{@{}ll@{}}
    \toprule
    Parameter & Search space\\
    \midrule
    $|C_a|$ & randint(0,100,10) \\
    $w$ & $\{0\} \cup$ 10**randint(-10,0)\\
    lr & loguniform($10^{-6},10^{-1}$) \\
    $\alpha$& 10**randint(0,10)\\
    $z$ & randint(10,100,10) \\
    $\gamma$ & 10**randint(0,10)\\
    layers & randint(1,3) \\
    filters & 2**randint(5,7)\\
    units & 2**randint(5,11)\\
    kernel size & [3, 5, 7]\\
    clipping & 10**randint(-10, 0) \\
    \bottomrule
    \end{tabular}
\end{table}
The batch size was fixed to \num{512}, meaning that each batch contained \num{512} labeled examples and the same amount of unlabeled examples.
In case of a size mismatch of labeled and unlabeled data, we re-sampled the smaller subset to fill the batches.
We used a cosine annealing learning rate scheduler and trained for \num{100} epochs.
Predictions on the test set were done using weights of the last epoch resulting in the best accuracy on the validation set (= \SI{20}{\percent} of the training set).
We increase $\lambda$ every epoch with a step size of $.1$, with a maximum of 1.

\section{Results \& Discussion}
\label{sec:discussion}
The results of our experiments described in~\cref{sec:experiments} can be seen in~\cref{tab:results}.
Some tables are available in the online appendix~\footnote{\url{https://github.com/LSX-UniWue/SuSL4TS}} as they only quantify the depicted observations.
\begin{table*}
\sisetup{
    round-mode          = places, 
    round-precision     = 2, 
}
    \caption{Results.
    Each block describes one dataset, and is subdivided with the baseline methods.
    For reference, we include the majority vote baseline, a fully supervised baseline,
    and a semi-unsupervised baseline with an MLP on the raw time series.
    For \algoo, we provide performance of our model in the different learning paradigms, with two versions of the semi-supervised and semi-unsupervised.
    We report the accuracy on the test set for unsupervised (UL), semi-supervised (SSL), semi-unsupervised (SuSL) and supervised (SL) learning paradigms.
    }
    \label{tab:results}
    \centering
    \begin{tabular}{@{}llSSSSSS@{}}
    \toprule
    \multirow{2}{*}{Dataset} & \multirow{2}{*}{Model} & {UL} & \multicolumn{2}{c}{SuSL} & \multicolumn{2}{c}{SSL} & {SL}\\
    \cmidrule{4-5}\cmidrule{6-7}
     & & \SI{0}{\percent} & \SI{20}{\percent} & \SI{50}{\percent} & \SI{20}{\percent} & \SI{50}{\percent} & \SI{100}{\percent}\\
    \midrule
    \multirow{7}{*}{HAR} & Majority Vote & & & & & & 18.22\\
    & Baseline (SVM,\cite{anguita2013public}) & & & & & & 96.4 \\
    & Baseline (MLP SuSL,\cite{willetts2018semi,willetts2020semi}) & 64.20 & & & 83.30 & 87.82 & 89.82\\
    \cmidrule{2-8}
    & \algoo & 65.38 & & & 92.70 & 91.72 & 92.33 \\
    & \algoo\ (movement (h)) & & 50.93 & 51.74 & & & \\
    & \algoo\ (stationary (h)) & & 87.98 & 87.81 & & & \\
    & \algoo\  (walking,laying (h)) & & 60.84 & 77.63 & & & \\
    \midrule
    \multirow{6}{*}{ECG} & Majority Vote & & & & & & 82.76\\
    & Baseline (CNN,\cite{Kachuee_2018}) & & & & & & 96.4 \\
    & Baseline (MLP SuSL,\cite{willetts2018semi,willetts2020semi}) & 84.56 & & & 96.94 & 97.60 & 98.21\\
    \cmidrule{2-8}
    & \algoo  & 83.28 & &  & 97.51 & 97.28 & 97.61\\
    & \algoo\  (q,v (h)) & &  84.70 & 84.25 & & & \\
    & \algoo\  (n (h)) & & 88.26 & 90.41 & & & \\
    \midrule
    \multirow{6}{*}{El. Devices} & Majority Vote & & & & & & 24.23\\
    & Baseline (BOSS,\cite{shifaz2020ts}) & & & & & & 79.92 \\
    & Baseline (MLP SuSL,\cite{willetts2018semi,willetts2020semi}) & 53.33 & & & 51.42 & 54.21 & 58.34\\
    \cmidrule{2-8}
    & \algoo & 52.72 & & & 69.69 & 68.02 & 69.96 \\
    & \algoo\ (1--3 (h)) &  & 54.59 & 59.14 & & & \\
    & \algoo\ (4--7 (h)) & & 49.51 & 59.61 & & & \\
    \bottomrule
    \end{tabular}
\end{table*}

\paragraph{Human Activity Recognition}
When taking a closer look at the HAR dataset, we see that our approach is not able to perform on par with the fully supervised baseline (SVM, 92 vs 96).
However, the SVM performs on extracted features of the signal, while we directly use the raw signal.
The main difference in performance can be attributed to the classes sitting and standing, as can be seen in~\cref{tab:confusion_har_fs}.
They are easily confused with each other.

But even with fewer labels (SSL, \SIlist{20;50}{\percent}), we achieve almost the same classification performance as with all labels (91 vs 92).
Surprisingly the accuracy is higher with fewer labeled samples (92 vs 93), which can be attributed to the better recall with the class standing.

In the unlabeled setting (i.e. time series clustering), the performance drops significantly (64 vs 92).
Furthermore, there is no difference between the fully connected approach and the convolutional one.

In the first semi-unsupervised setting (walking classes hidden), we can see worse performance in both (\SIlist{20;50}{\percent}) settings than compared to the unsupervised clustering.
This drop in accuracy is due to the fact that all walking related classes are classified as walking, and even standing is classified as walking (see~\cref{tab:confusion_har_susl_walking}).
In contrast to the unsupervised model within the semi-unsupervised task, both parts of the encoder, the labeled and unlabeled leg, have to be trained, implying more weights need to be tuned for the network with only few labeled samples left.
This setting is thus more complicated than the unsupervised task and the models tend to group unknown classes into one unknown super-class.

On the other hand, when hiding all stationary classes (i.e. standing, laying and sitting), performance increases again (51 vs 88).
Wrongfully assigned classes are mostly confusion of sitting and standing (see~\cref{tab:confusion_har_susl_static}).

In the last semi-unsupervised setting (hiding walking and laying), performance is a little lower than hiding all static classes (77 vs 88).
The drop in performance is due to the fact that the class standing is completely missed and most samples are classified as walking (see~\cref{tab:confusion_har_susl_mixed}).

\begin{figure*}
\centering
\begin{subfigure}[t]{0.3\textwidth}
    \includegraphics[width=\textwidth]{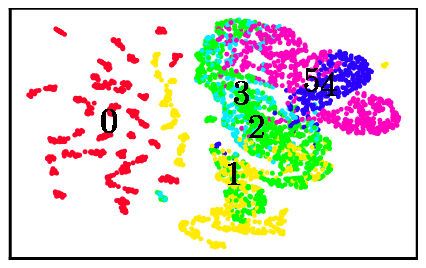}
    \caption{Unsupervised classification.}
    \label{fig:results_embedding_har_us}
\end{subfigure}
\hfill
\begin{subfigure}[t]{0.3\textwidth}
    \includegraphics[width=\textwidth]{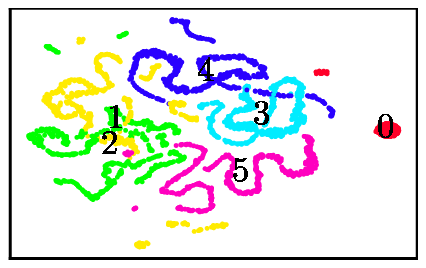}
    \caption{Semi-supervised classification with \SI{20}{\percent} labels.}
    \label{fig:results_embedding_har_ss}
\end{subfigure}
\hfill
\begin{subfigure}[t]{0.3\textwidth}
    \includegraphics[width=\textwidth]{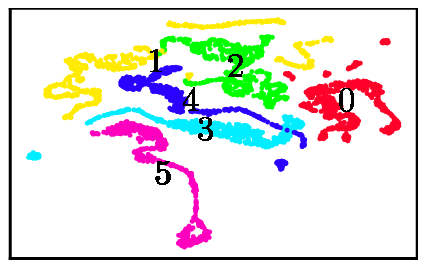}
    \caption{Fully supervised classification.}
    \label{fig:results_embedding_har_fs}
\end{subfigure}
\hfill
\begin{subfigure}[t]{\textwidth}
    \includegraphics[width=\textwidth]{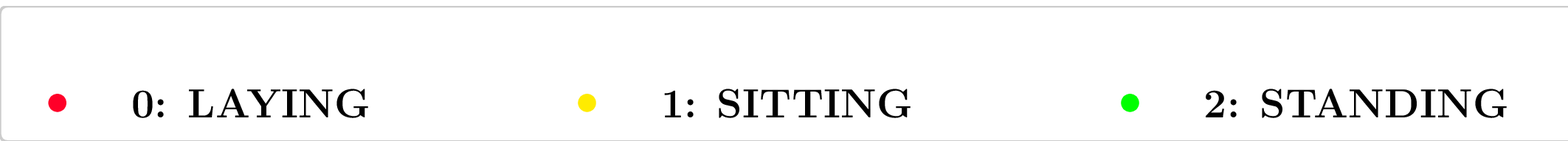}
\end{subfigure}
\hfill
\begin{subfigure}[t]{0.3\textwidth}
    \includegraphics[width=\textwidth]{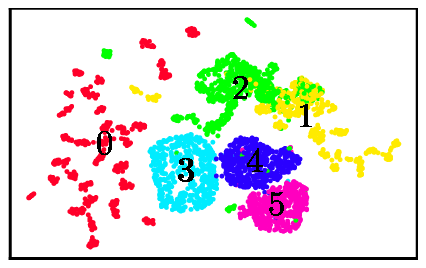}
    \caption{Semi-unsupervised classification with \SI{20}{\percent} and the hidden classes of Sitting (1), Standing (2) and Laying (0).}
    \label{fig:results_embedding_har_susl_static}
\end{subfigure}
\hfill
\begin{subfigure}[t]{0.3\textwidth}
    \includegraphics[width=\textwidth]{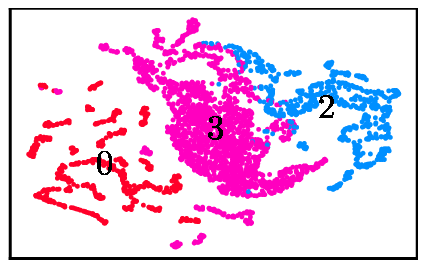}
    \caption{Semi-unsupervised classification with \SI{20}{\percent} and the hidden classes of Walking (3), Walking Downstairs (4) and Walking Upstairs (5).}
    \label{fig:results_embedding_har_susl_movement}
\end{subfigure}
\hfill
\begin{subfigure}[t]{0.3\textwidth}
    \includegraphics[width=\textwidth]{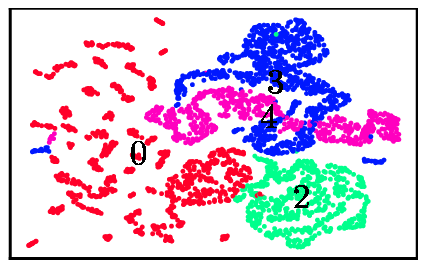}
    \caption{Semi-unsupervised classification with \SI{20}{\percent} and the hidden classes of Laying (0) and Walking (3).}
    \label{fig:results_embedding_har_susl_mixed}
\end{subfigure}
    \caption{Embedding visualization.
    UMAP dimensionality reduction of the learned latent space on the test set for the HAR dataset.}
    \label{fig:results_embedding}
\end{figure*}

All discussed observations for the HAR dataset are also visible in the visualization of the latent space shown in~\cref{fig:results_embedding}.
We used UMAP~\cite{mcinnes2018umap}(min\_dist=0.99, n\_neighbors=10, metric=cosine) to plot a two dimensional manifold of the learned embedding.

\paragraph{ECG Heartbeat Classification}
The classification results for the univariate ECG dataset are displayed in~\cref{tab:results}.
When comparing with the fully supervised CNN architecture presented in~\cite{Kachuee_2018}, both semi-unsupervised approaches outperform the baseline (98 vs 96).

In the semi-supervised setting (\SIlist{20;50}{\percent}, all classes known), there is almost no difference in terms of accuracy compared to the supervised settings.
Again, accuracies are slightly higher when using fewer labels.
Compared to the fully connected SuSL model, the convolutional feature extractor fares slightly better (97 vs 98).

In the unsupervised setting we can observe a performance drop, although not as high as for the HAR dataset (83 vs 63).
Due to the highly skewed nature of the dataset, the unsupervised classification task is not much better than the majority vote (\SI{82.8}{\percent}), suggesting that only the normal class was detected (see~\cref{tab:confusion_ecg_ul}).

In the first semi-unsupervised setting (classes Q and V hidden), we can see a slight increase in performance in comparison to the unsupervised setting (84 vs 85).
Both hidden classes are missed in the test set classification, where the differences in accuracies are founded in better recall of classes F and S (see~\cref{tab:confusion_ecg_susl_02_qv,tab:confusion_ecg_susl_05_qv}).

On the other hand, when hiding the normal class, performance increases above the majority vote (88 vs 83).
The amount of available labels (\SIlist{20;50}{\percent}) does not impact the predictions largely (88 vs 90).
However, with \SI{50}{\percent} available labels, the class F is not missed, class V is missed in both settings (see~\cref{tab:confusion_ecg_susl_02_n,tab:confusion_ecg_susl_05_n}).

\paragraph{Electric Devices}
The classification results for the univariate electric devices dataset are displayed in~\cref{tab:results}.
When comparing with the best performing model (BOSS) in the supervised settings, both neural network approaches perform worse (80 vs 70), where most mis-classification are done within the classes 3--5 (see~\cref{tab:confusion_ed_sl}).

In the semi-supervised setting (\SIlist{20;50}{\percent}, all classes known), we observe the same behavior as in the ECG datasets, classification accuracy remains similar to the fully supervised setting (68 vs 70).
Since class 7 is completely missed (\SI{20}{\percent}), the model likely overfitted, since this class is detected in the validation set (see~\cref{tab:confusion_ed_ssl_02,tab:confusion_ed_ssl_05}).

In the unsupervised classification task, we can observe that two classes are missed in the test set (1, 7), but even predictions within the other classes are not very clear (52 vs 70).
The missed classes only represent smaller portions of this dataset, suggesting the parameter search found models yielding higher accuracy when predicting mostly the larger classes (see~\cref{tab:confusion_ed_ul}).

In the first semi-unsupervised setting (classes 1--3 hidden), we observe a larger dip in accuracy compared to the semi-supervised setting (55/59 vs 70).
Within this dataset we can see an increase in performance when comparing the setting with \SI{20}{\percent} available labels, in contrast to the \SI{50}{\percent} configuration (55 vs 59, see~\cref{tab:confusion_ed_susl_02_1-3,tab:confusion_ed_susl_05_1-3}).
In both settings, one class is completely missed (1 or 7), but the main difference in performance is the higher precision at all other classes with \SI{50}{\percent} labels available.

On the other hand, when hiding classes 4--7, we can observe a similar decline in performance (50/60 vs 70).
With \SI{20}{\percent} labels available, the model assigns mostly all missed samples to the same hidden class (5), while missing classes 1, 4 and 6--7 completely (see~\cref{tab:confusion_ed_susl_02_4-7}).
When presenting the model more labels, accuracy once again increases (similar to classes 1--3 hidden), and only class 1 is completely missed (see~\cref{tab:confusion_ed_susl_05_4-7}).

\paragraph{General Discussion}
Throughout all datasets and settings, we have made some observations applicable in general, which we will discuss now.

(1) In the multivariate dataset (i.e. HAR), the convolutional approach of \algoo\ outperforms the fully connected version for semi-unsupervised learning.
That is, it performs better in any setting we tested.

(2) In the univariate datasets, we can see a mixed picture.
For the electric devices dataset, \algoo\ performs better in any labeled setting by a large margin.
Given the ECG dataset, both versions perform equally well, with no clear tendency.

(3) If specific classes are not known, we can see drastically different results in the classification.
Most prominent in the HAR dataset, as hiding the walking classes collapses predictions to perform worse than the unsupervised setting.

(4) In general the semi-unsupervised setting, only when using the larger amount of \SI{50}{\percent} labels available, we can see an increased performance compared to the unsupervised settings with no labels at all (except HAR with all movements hidden).

(5) Generally the semi-supervised (i.e. all classes known, only limited amount of labels) performance is as good as the fully supervised setting.

\section{Summary}
\label{sec:summary}
In this paper we presented \algoo, a convolutional Gaussian mixture model for performing semi-unsupervised time series classification.
We showed the efficacy of our approach by comparing it with optimized methods on several benchmark datasets, while requiring no manual feature extraction.
Especially in the semi-supervised settings, the model performs nearly as good as its fully supervised counterpart.
When omitting specific classes (i.e. classes unknown a priori)
, accuracy can highly deviate in certain combinations of labeled versus unlabeled data, showing lower performance than using no labels at all.

In future work, we will analyze the applicability of our approach in real world, large scale data.
For example, we could test the highly skewed sensor data obtained from beehives~\cite{zacepins2016remote,kviesis2016application,davidson2020anomaly,davidson2021anomaly}, or other highly skewed anomaly detection datasets~\cite{ren2019time}, alleviating the burden of having to manually discern the different types of anomalies.
On the other hand, we could use the normal class completely unlabeled and only annotate a few anomaly classes.
This dataset is similar to the presented ECG analysis.
In a more complex setting of time series classification, we could try to classify audio files, either with extracted mel-spectrograms or the raw series~\cite{kahl2020overview,piczak2015esc,janetzky2021detecting}.

As mentioned, \algoo\ is a generative model, thus we can draw random samples resembling the learned classes from the latent space.
That enables us to generate samples for a given class to be used for other tasks or augment the labeled set in the whole dataset.

\bibliographystyle{ACM-Reference-Format}
\bibliography{bibliography}

\section*{Appendix}
\label{sec:appendix}
\balance
\begin{table}[H]
    \centering
\begin{subtable}{\linewidth}
    \centering
    \caption{Class distribution in the HAR dataset}
    \label{tab:har_classes}
    \begin{tabular}{@{}lSSS@{}}
        \toprule
        Class & {Training samples} & {Test samples} & {$\Sigma$} \\
        \midrule
        Walking & 1226 & 496 & 1722 \\
        Walking up & 1073 & 471 & 1544 \\
        Walking down & 986 & 420 & 1406 \\
        Sitting & 1286 & 491 & 1777 \\
        Standing & 1374 & 532 & 1906 \\
        Laying & 1407 & 537 & 1944 \\
        \midrule
        & 7352 & 2947 & 10299 \\
        \bottomrule
    \end{tabular}
\end{subtable}
\hfill
\begin{subtable}{\linewidth}
    \centering
    \caption{Class distribution in the MIT-BIH dataset}
    \label{tab:ecg_classes}
    \begin{tabular}{@{}lSSS@{}}
        \toprule
        Class & {Training samples} & {Test samples} & {$\Sigma$} \\
        \midrule
        N & 72471 & 18118 & 90589 \\
        S & 2223 & 557 & 2780 \\
        V & 5788 & 1448 & 7236 \\
        F & 641 & 162 & 803 \\
        Q & 6431 & 1607 & 8038 \\
        \midrule
        & 87554 & 21892 & 109446 \\
        \bottomrule
    \end{tabular}
\end{subtable}
\hfill
\begin{subtable}{\linewidth}
    \centering
    \caption{Class distribution in the electric devices dataset}
    \label{tab:eld_class}
    \begin{tabular}{@{}lSSS@{}}
        \toprule
        Class & {Training samples} & {Test samples} & {$\Sigma$} \\
        \midrule
        1 & 727 & 667 & 1394 \\
        2 & 2231 & 1956 & 4187 \\
        3 & 851 & 755 & 1606 \\
        4 & 1474 & 1165 & 2639 \\
        5 & 2406 & 1869 & 4275 \\
        6 & 509 & 743 & 1252 \\
        7 & 728 & 556 & 1284 \\
        \midrule
        & 8926 & 7711 & 16637 \\
        \bottomrule
    \end{tabular}
\end{subtable}
\end{table}

\clearpage
\begin{table*}
    \centering
\begin{subtable}{\textwidth}
\sisetup{
    round-mode          = places, 
    round-precision     = 2, 
}
    \caption{Macro F1 Scores
    }
    \label{tab:macro_f1}
    \centering
    \begin{tabular}{@{}llSSSSSS@{}}
    \toprule
    \multirow{2}{*}{Dataset} & \multirow{2}{*}{Model} & {UL} & \multicolumn{2}{c}{SuSL} & \multicolumn{2}{c}{SSL} & {SL}\\
    \cmidrule{4-5}\cmidrule{6-7}
     & & \SI{0}{\percent} & \SI{20}{\percent} & \SI{50}{\percent} & \SI{20}{\percent} & \SI{50}{\percent} & \SI{100}{\percent}\\
    \midrule
    \multirow{5}{*}{HAR}
    & Baseline (MLP SuSL,\cite{willetts2018semi,willetts2020semi}) & 61.96 & & & 83.01 & 87.85 & 89.79\\
    \cmidrule{2-8}
    & \algoo & 64.46 & & & 92.73 & 91.80 & 93.28 \\
    & \algoo\ (movement (h)) & & 37.64 & 38.69 & & & \\
    & \algoo\ (stationary (h)) & & 79.32 & 87.95 & & & \\
    & \algoo\  (walking,laying (h)) & & 49.59 & 73.33 & & & \\
    \midrule
    \multirow{4}{*}{ECG}
    & Baseline (MLP SuSL,\cite{willetts2018semi,willetts2020semi}) & 32.22 & & & 84.30 & 87.23 & 90.55\\
    \cmidrule{2-8}
    & \algoo  & 28.32 & &  & 86.68 & 85.66 & 86.34\\
    & \algoo\  (q,v (h)) & &  46.04 & 41.21 & & & \\
    & \algoo\  (n (h)) & & 46.43 & 53.32 & & & \\
    \midrule
    \multirow{4}{*}{El. Devices}
    & Baseline (MLP SuSL,\cite{willetts2018semi,willetts2020semi}) & 43.59 & & & 40.09 & 39.59 & 45.93\\
    \cmidrule{2-8}
    & \algoo & 37.02 & & & 61.91 & 61.18 & 57.45 \\
    & \algoo\ (1--3 (h)) &  & 42.25 & 44.14 & & & \\
    & \algoo\ (4--7 (h)) & & 27.35 & 38.08 & & & \\
    \bottomrule
    \end{tabular}
\end{subtable}
\hfill
\begin{subtable}{\textwidth}
\sisetup{
    round-mode          = places, 
    round-precision     = 2, 
}
    \caption{Weighted F1 Scores
    }
    \label{tab:weighted_f1}
    \centering
    \begin{tabular}{@{}llSSSSSS@{}}
    \toprule
    \multirow{2}{*}{Dataset} & \multirow{2}{*}{Model} & {UL} & \multicolumn{2}{c}{SuSL} & \multicolumn{2}{c}{SSL} & {SL}\\
    \cmidrule{4-5}\cmidrule{6-7}
     & & \SI{0}{\percent} & \SI{20}{\percent} & \SI{50}{\percent} & \SI{20}{\percent} & \SI{50}{\percent} & \SI{100}{\percent}\\
    \midrule
    \multirow{5}{*}{HAR}
    & Baseline (MLP SuSL,\cite{willetts2018semi,willetts2020semi}) & 65.48 & & & 83.26 & 87.72 & 89.82\\
    \cmidrule{2-8}
    & \algoo & 65.34 & & & 92.76 & 91.73 & 93.21 \\
    & \algoo\ (movement (h)) & & 61.44 & 61.92 & & & \\
    & \algoo\ (stationary (h)) & & 87.83 & 87.89 & & & \\
    & \algoo\  (walking,laying (h)) & & 71.87 & 83.34 & & & \\
    \midrule
    \multirow{4}{*}{ECG}
    & Baseline (MLP SuSL,\cite{willetts2018semi,willetts2020semi}) & 89.03 & & & 97.00 & 97.70 & 98.24\\
    \cmidrule{2-8}
    & \algoo  & 87.55 & &  & 97.66 & 97.39 & 97.77\\
    & \algoo\  (q,v (h)) & &  91.14 & 90.69 & & & \\
    & \algoo\  (n (h)) & & 91.82 & 93.98 & & & \\
    \midrule
    \multirow{4}{*}{El. Devices}
    & Baseline (MLP SuSL,\cite{willetts2018semi,willetts2020semi}) & 55.92 & & & 54.57 & 59.33 & 61.35\\
    \cmidrule{2-8}
    & \algoo & 58.15 & & & 72.83 & 68.42 & 71.10 \\
    & \algoo\ (1--3 (h)) &  & 59.22 & 63.34 & & & \\
    & \algoo\ (4--7 (h)) & & 61.71 & 64.77 & & & \\
    \bottomrule
    \end{tabular}
\end{subtable}
\end{table*}
\clearpage
\begin{table*}
    \centering
\begin{subtable}{0.45\textwidth}
	\centering
	\caption{Confusion Matrix HAR UL\\
	Accuracy: 0.6538853070919579}
	\label{tab:confusion_har_us}
	\begin{tabular}{@{}SSSSSSS@{}}
		\toprule
		{L} & {Si} & {St} & {W} & {W Down} & {W Up} \\
		\midrule
		494 & 0 & 1 & 13 & 23 & 6 \\
		4 & 349 & 118 & 4 & 12 & 4 \\
		0 & 86 & 387 & 22 & 14 & 23 \\
		0 & 22 & 85 & 280 & 49 & 60 \\
		0 & 7 & 17 & 83 & 213 & 100 \\
		0 & 5 & 42 & 92 & 128 & 204 \\
		\bottomrule
	\end{tabular}
\end{subtable}
\hfill
\begin{subtable}{0.45\textwidth}
	\centering
	\caption{Confusion Matrix HAR SUSL \SI{20}{\percent} (movement hidden)\\
	Accuracy: 0.509331523583305}
	\label{tab:confusion_har_susl_walking}
	\begin{tabular}{@{}SSSSSSS@{}}
		\toprule
		{L} & {Si} & {St} & {W (h)} & {W Down (h)} & {W Up (h)} \\
		\midrule
		528 & 0 & 0 & 9 & 0 & 0 \\
		0 & 0 & 128 & 363 & 0 & 0 \\
		0 & 0 & 479 & 53 & 0 & 0 \\
		0 & 0 & 2 & 494 & 0 & 0 \\
		0 & 0 & 0 & 420 & 0 & 0 \\
		0 & 0 & 1 & 470 & 0 & 0 \\
		\bottomrule
	\end{tabular}
\end{subtable}
\hfill
\begin{subtable}{0.45\textwidth}
	\centering
	\caption{Confusion Matrix HAR SUSL \SI{20}{\percent}  (stationary hidden)\\
	Accuracy: 0.8798778418730913}
	\label{tab:confusion_har_susl_static}
	\begin{tabular}{@{}SSSSSSS@{}}
		\toprule
		{L (h)} & {Si (h)} & {St (h)} & {W} & {W Down} & {W Up} \\
		\midrule
		494 & 0 & 43 & 0 & 0 & 0 \\
		0 & 360 & 123 & 0 & 2 & 6 \\
		0 & 64 & 465 & 1 & 1 & 1 \\
		0 & 0 & 0 & 449 & 46 & 1 \\
		0 & 0 & 0 & 4 & 408 & 8 \\
		0 & 0 & 2 & 5 & 47 & 417 \\
		\bottomrule
	\end{tabular}
\end{subtable} \hfill
\begin{subtable}{0.45\textwidth}
	\centering
	\caption{Confusion Matrix HAR SUSL \SI{50}{\percent} hide on each\\
	Accuracy: 0.7763827621309807}
	\label{tab:confusion_har_susl_mixed}
	\begin{tabular}{@{}SSSSSSS@{}}
		\toprule
		{L (h)} & {Si} & {St} & {W (h)} & {W Down} & {W Up} \\
		\midrule
		537 & 0 & 0 & 0 & 0 & 0 \\
		1 & 400 & 0 & 87 & 0 & 3 \\
		0 & 78 & 0 & 454 & 0 & 0 \\
		1 & 0 & 0 & 474 & 0 & 21 \\
		0 & 0 & 0 & 0 & 417 & 3 \\
		0 & 0 & 0 & 7 & 4 & 460 \\
		\bottomrule
	\end{tabular}
\end{subtable} \hfill
\begin{subtable}{0.45\textwidth}
	\centering
	\caption{Confusion Matrix HAR SSL \SI{20}{\percent}\\
	Accuracy: 0.9270444519850696}
	\label{tab:confusion_har_ss}
	\begin{tabular}{@{}SSSSSSS@{}}
		\toprule
		{L} & {Si} & {St} & {W} & {W Down} & {W Up} \\
		\midrule
		537 & 0 & 0 & 0 & 0 & 0 \\
		3 & 393 & 72 & 0 & 0 & 23 \\
		0 & 78 & 450 & 1 & 0 & 3 \\
		0 & 0 & 0 & 485 & 9 & 2 \\
		0 & 0 & 0 & 2 & 413 & 5 \\
		0 & 2 & 1 & 1 & 13 & 454 \\
		\bottomrule
	\end{tabular}
\end{subtable} \hfill
\begin{subtable}{0.45\textwidth}
	\centering
	\caption{Confusion Matrix HAR SL\\
	Accuracy: 0.9317950458092976}
	\label{tab:confusion_har_fs}
	\begin{tabular}{@{}SSSSSSS@{}}
		\toprule
		{L} & {Si} & {St} & {W} & {W Down} & {W Up} \\
		\midrule
		537 & 0 & 0 & 0 & 0 & 0 \\
		0 & 416 & 49 & 1 & 0 & 25 \\
		0 & 94 & 437 & 1 & 0 & 0 \\
		0 & 0 & 1 & 492 & 3 & 0 \\
		0 & 0 & 0 & 3 & 417 & 0 \\
		0 & 0 & 0 & 24 & 0 & 447 \\
		\bottomrule
	\end{tabular}
\end{subtable}
\end{table*}

\clearpage
\begin{table*}
    \centering
\begin{subtable}{0.45\textwidth}
	\centering
	\caption{Confusion Matrix El. Devices UL \\
	Accuracy: 0.5272986642458825}
	\label{tab:confusion_ed_ul}
	\begin{tabular}{@{}SSSSSSS@{}}
		\toprule
		{1} & {2} & {3} & {4} & {5} & {6} & {7} \\
		\midrule
		0 & 225 & 51 & 151 & 137 & 103 & 0 \\
		0 & 1780 & 44 & 71 & 43 & 18 & 0 \\
		0 & 101 & 380 & 124 & 32 & 118 & 0 \\
		0 & 71 & 19 & 824 & 165 & 86 & 0 \\
		0 & 618 & 76 & 198 & 853 & 124 & 0 \\
		0 & 78 & 195 & 173 & 68 & 229 & 0 \\
		0 & 55 & 57 & 231 & 59 & 154 & 0 \\
		\bottomrule
	\end{tabular}
\end{subtable} \hfill
\begin{subtable}{0.45\textwidth}
	\centering
	\caption{Confusion Matrix El. Devices SUSL \SI{20}{\percent} (1-3 hidden) \\
	Accuracy: 0.5459732849176501}
	\label{tab:confusion_ed_susl_02_1-3}
	\begin{tabular}{@{}SSSSSSS@{}}
		\toprule
		{1 (h)} & {2 (h)} & {3 (h)} & {4} & {5} & {6} & {7} \\
		\midrule
		0 & 230 & 17 & 141 & 116 & 161 & 2 \\
		0 & 1813 & 38 & 59 & 33 & 12 & 1 \\
		0 & 192 & 389 & 85 & 15 & 66 & 8 \\
		0 & 67 & 31 & 852 & 117 & 91 & 7 \\
		0 & 662 & 43 & 248 & 779 & 88 & 49 \\
		0 & 88 & 89 & 163 & 61 & 300 & 42 \\
		0 & 55 & 23 & 273 & 46 & 82 & 77 \\
		\bottomrule
	\end{tabular}
\end{subtable} \hfill
\begin{subtable}{0.45\textwidth}
	\centering
	\caption{Confusion Matrix El. Devices SUSL \SI{50}{\percent} (1-3 hidden) \\
	Accuracy: 0.5914926728050837}
	\label{tab:confusion_ed_susl_05_1-3}
	\begin{tabular}{@{}SSSSSSS@{}}
		\toprule
		{1 (h)} & {2 (h)} & {3 (h)} & {4} & {5} & {6} & {7} \\
		\midrule
		78 & 174 & 37 & 230 & 64 & 84 & 0 \\
		0 & 1623 & 9 & 150 & 174 & 0 & 0 \\
		1 & 329 & 277 & 129 & 2 & 17 & 0 \\
		23 & 36 & 15 & 956 & 83 & 52 & 0 \\
		11 & 217 & 58 & 211 & 1359 & 13 & 0 \\
		16 & 115 & 45 & 255 & 44 & 268 & 0 \\
		2 & 16 & 148 & 306 & 81 & 3 & 0 \\
		\bottomrule
	\end{tabular}
\end{subtable} \hfill
\begin{subtable}{0.45\textwidth}
	\centering
	\caption{Confusion Matrix El. Devices SUSL \SI{20}{\percent} (4-7 hidden) \\
	Accuracy: 0.49513681753339384}
	\label{tab:confusion_ed_susl_02_4-7}
	\begin{tabular}{@{}SSSSSSS@{}}
		\toprule
		{1} & {2} & {3} & {4 (h)} & {5 (h)} & {6 (h)} & {7 (h)} \\
		\midrule
		0 & 193 & 21 & 0 & 453 & 0 & 0 \\
		0 & 1936 & 7 & 0 & 13 & 0 & 0 \\
		0 & 7 & 723 & 0 & 25 & 0 & 0 \\
		0 & 39 & 214 & 0 & 912 & 0 & 0 \\
		0 & 677 & 33 & 0 & 1159 & 0 & 0 \\
		0 & 77 & 169 & 0 & 497 & 0 & 0 \\
		0 & 66 & 169 & 0 & 321 & 0 & 0 \\
		\bottomrule
	\end{tabular}
\end{subtable} \hfill
\begin{subtable}{0.45\textwidth}
	\centering
	\caption{Confusion Matrix El. Devices SUSL \SI{50}{\percent} (4-7 hidden) \\
	Accuracy: 0.5961613279730256}
	\label{tab:confusion_ed_susl_05_4-7}
	\begin{tabular}{@{}SSSSSSS@{}}
		\toprule
		{1} & {2} & {3} & {4 (h)} & {5 (h)} & {6 (h)} & {7 (h)} \\
		\midrule
		0 & 192 & 136 & 144 & 157 & 32 & 6 \\
		0 & 1868 & 28 & 16 & 29 & 2 & 13 \\
		0 & 31 & 571 & 57 & 72 & 11 & 13 \\
		0 & 45 & 75 & 829 & 135 & 54 & 27 \\
		0 & 599 & 71 & 123 & 1000 & 23 & 53 \\
		0 & 75 & 373 & 128 & 77 & 74 & 16 \\
		0 & 28 & 74 & 67 & 125 & 7 & 255 \\
		\bottomrule
	\end{tabular}
\end{subtable} \hfill
\begin{subtable}{0.45\textwidth}
	\centering
	\caption{Confusion Matrix El. Devices SSL \SI{20}{\percent} \\
	Accuracy: 0.6969264686811049}
	\label{tab:confusion_ed_ssl_02}
	\begin{tabular}{@{}SSSSSSS@{}}
		\toprule
		{1} & {2} & {3} & {4} & {5} & {6} & {7} \\
		\midrule
		233 & 7 & 8 & 118 & 195 & 106 & 0 \\
		0 & 1586 & 6 & 50 & 314 & 0 & 0 \\
		12 & 0 & 706 & 3 & 9 & 25 & 0 \\
		99 & 0 & 124 & 783 & 112 & 47 & 0 \\
		9 & 90 & 24 & 86 & 1655 & 5 & 0 \\
		109 & 1 & 27 & 123 & 72 & 411 & 0 \\
		7 & 98 & 155 & 40 & 255 & 1 & 0 \\
		\bottomrule
	\end{tabular}
\end{subtable} \hfill
\begin{subtable}{0.45\textwidth}
	\centering
	\caption{Confusion Matrix El. Devices SSL \SI{50}{\percent} \\
	Accuracy: 0.6801971209959797}
	\label{tab:confusion_ed_ssl_05}
	\begin{tabular}{@{}SSSSSSS@{}}
		\toprule
		{1} & {2} & {3} & {4} & {5} & {6} & {7} \\
		\midrule
		217 & 1 & 4 & 84 & 223 & 136 & 2 \\
		0 & 1450 & 2 & 39 & 460 & 0 & 5 \\
		9 & 1 & 528 & 17 & 8 & 18 & 174 \\
		106 & 0 & 100 & 693 & 148 & 93 & 25 \\
		6 & 98 & 10 & 75 & 1640 & 9 & 31 \\
		42 & 0 & 21 & 90 & 65 & 503 & 22 \\
		10 & 21 & 1 & 43 & 266 & 1 & 214 \\
		\bottomrule
	\end{tabular}
\end{subtable} \hfill
\begin{subtable}{0.45\textwidth}
	\centering
	\caption{Confusion Matrix El. Devices SL \\
	Accuracy: 0.6996498508624044}
	\label{tab:confusion_ed_sl}
	\begin{tabular}{@{}SSSSSSS@{}}
		\toprule
		{1} & {2} & {3} & {4} & {5} & {6} & {7} \\
		\midrule
		149 & 1 & 15 & 129 & 210 & 161 & 2 \\
		0 & 1533 & 1 & 23 & 395 & 0 & 4 \\
		6 & 0 & 569 & 9 & 1 & 13 & 157 \\
		49 & 0 & 148 & 758 & 116 & 78 & 16 \\
		11 & 77 & 14 & 52 & 1692 & 2 & 21 \\
		35 & 0 & 25 & 108 & 52 & 504 & 19 \\
		7 & 11 & 2 & 46 & 298 & 2 & 190 \\
		\bottomrule
	\end{tabular}
\end{subtable}
\end{table*}
\clearpage
\begin{table*}
    \centering
\begin{subtable}{0.45\textwidth}
	\centering
	\caption{Confusion Matrix ECG UL \\
	Accuracy: 0.8328613192033619}
	\label{tab:confusion_ecg_ul}
	\begin{tabular}{SSSSS}
		\toprule
		{N} & {S} & {V} & {F} & {Q} \\
		\midrule
		17354 & 0 & 0 & 0 & 764 \\
		541 & 0 & 0 & 0 & 15 \\
		1243 & 0 & 0 & 0 & 205 \\
		162 & 0 & 0 & 0 & 0 \\
		729 & 0 & 0 & 0 & 879 \\
		\bottomrule
	\end{tabular}
\end{subtable} \hfill
\begin{subtable}{0.45\textwidth}
	\centering
	\caption{Confusion Matrix ECG SUSL \SI{20}{\percent} (Q,V hidden) \\
	Accuracy: 0.8470217431025032}
	\label{tab:confusion_ecg_susl_02_qv}
	\begin{tabular}{SSSSS}
		\toprule
		{N} & {S} & {V (h)} & {F} & {Q (h)} \\
		\midrule
		18070 & 43 & 0 & 5 & 0 \\
		202 & 354 & 0 & 0 & 0 \\
		1340 & 40 & 0 & 68 & 0 \\
		43 & 0 & 0 & 119 & 0 \\
		1607 & 1 & 0 & 0 & 0 \\
		\bottomrule
	\end{tabular}
\end{subtable} \hfill
\begin{subtable}{0.45\textwidth}
	\centering
	\caption{Confusion Matrix ECG SUSL \SI{50}{\percent} (Q,V hidden) \\
	Accuracy: 0.8425452219989037}
	\label{tab:confusion_ecg_susl_05_qv}
	\begin{tabular}{SSSSS}
		\toprule
		{N} & {S} & {V (h)} & {F} & {Q (h)} \\
		\midrule
		18050 & 55 & 0 & 13 & 0 \\
		260 & 296 & 0 & 0 & 0 \\
		1233 & 151 & 0 & 64 & 0 \\
		61 & 2 & 0 & 99 & 0 \\
		1605 & 3 & 0 & 0 & 0 \\
		\bottomrule
	\end{tabular}
\end{subtable} \hfill
\begin{subtable}{0.45\textwidth}
	\centering
	\caption{Confusion Matrix ECG SUSL \SI{20}{\percent} (N hidden) \\
	Accuracy: 0.8826511967842134}
	\label{tab:confusion_ecg_susl_02_n}
	\begin{tabular}{SSSSS}
		\toprule
		{N (h)} & {S} & {V} & {F} & {Q} \\
		\midrule
		17634 & 51 & 0 & 0 & 433 \\
		308 & 245 & 0 & 0 & 3 \\
		1415 & 15 & 0 & 0 & 18 \\
		161 & 0 & 0 & 0 & 1 \\
		164 & 0 & 0 & 0 & 1444 \\
		\bottomrule
	\end{tabular}
\end{subtable} \hfill
\begin{subtable}{0.45\textwidth}
	\centering
	\caption{Confusion Matrix ECG SUSL \SI{50}{\percent} (N hidden) \\
	Accuracy: 0.9041659053535538}
	\label{tab:confusion_ecg_susl_05_n}
	\begin{tabular}{SSSSS}
		\toprule
		{N (h)} & {S} & {V} & {F} & {Q} \\
		\midrule
		17992 & 44 & 0 & 0 & 82 \\
		287 & 269 & 0 & 0 & 0 \\
		1437 & 1 & 0 & 0 & 10 \\
		148 & 0 & 0 & 14 & 0 \\
		89 & 0 & 0 & 0 & 1519 \\
		\bottomrule
	\end{tabular}
\end{subtable} \hfill
\begin{subtable}{0.45\textwidth}
	\centering
	\caption{Confusion Matrix ECG SSL \SI{20}{\percent} \\
	Accuracy: 0.9751964187831171}
	\label{tab:confusion_ecg_ss_02}
	\begin{tabular}{SSSSS}
		\toprule
		{N} & {S} & {V} & {F} & {Q} \\
		\midrule
		18055 & 23 & 32 & 5 & 3 \\
		203 & 344 & 9 & 0 & 0 \\
		108 & 2 & 1322 & 15 & 1 \\
		46 & 0 & 18 & 98 & 0 \\
		64 & 0 & 14 & 0 & 1530 \\
		\bottomrule
	\end{tabular}
\end{subtable} \hfill
\begin{subtable}{0.45\textwidth}
	\centering
	\caption{Confusion Matrix ECG SSL \SI{50}{\percent} \\
	Accuracy: 0.9727754430842317}
	\label{tab:confusion_ecg_ss_05}
	\begin{tabular}{SSSSS}
		\toprule
		{N} & {S} & {V} & {F} & {Q} \\
		\midrule
		17988 & 44 & 58 & 15 & 13 \\
		194 & 345 & 14 & 0 & 3 \\
		87 & 4 & 1328 & 21 & 8 \\
		45 & 1 & 13 & 103 & 0 \\
		63 & 4 & 7 & 2 & 1532 \\
		\bottomrule
	\end{tabular}
\end{subtable} \hfill
\begin{subtable}{0.45\textwidth}
	\centering
	\caption{Confusion Matrix ECG SL \\
	Accuracy: 0.9761099945185456}
	\label{tab:confusion_ecg_fs}
	\begin{tabular}{SSSSS}
		\toprule
		{N} & {S} & {V} & {F} & {Q} \\
		\midrule
		18078 & 10 & 22 & 4 & 4 \\
		217 & 330 & 8 & 0 & 1 \\
		111 & 5 & 1305 & 20 & 7 \\
		51 & 0 & 14 & 97 & 0 \\
		42 & 0 & 7 & 0 & 1559 \\
		\bottomrule
	\end{tabular}
\end{subtable}
\end{table*}

\end{document}